\documentclass[letterpaper, 10 pt, conference]{ieeeconf}  

\IEEEoverridecommandlockouts                              

\usepackage{graphicx} 
\usepackage{booktabs}
\usepackage{multirow}
\usepackage{amsmath, amsfonts, amssymb}
\usepackage{algorithm}
\usepackage{algpseudocode}
\usepackage{hyperref}

\title{\LARGE \bf
Vessel Traffic Flow Prediction on Sparse Data via Spatio-Temporal Graph Neural Networks with a Learnable Tweedie Head
}

\author{Kyeongjun  Lee$^{a}$ and Heeyoung Kim$^{a,\dagger}$%
  \thanks{$^{a}$Department of Industrial and Systems Engineering,
    Korea Advanced Institute of Science and Technology (KAIST),
    Daejeon, Republic of Korea.}%
  \thanks{$^\dagger$Corresponding author:
    \href{mailto:heeyoungkim@kaist.ac.kr}{heeyoungkim@kaist.ac.kr}}%
}

\begin{document}

\maketitle
\thispagestyle{empty}
\pagestyle{empty}

\begin{abstract}
Accurate vessel traffic flow prediction is crucial for smart port operations and navigational safety. 
However, maritime traffic flow data are often highly sparse with intermittent bursts, making robust forecasting challenging. 
Under such conditions, conventional spatio-temporal graph neural networks (ST-GNNs) can degrade toward conservative near-zero predictions and fail to capture non-zero activity. 
Although zero-inflated negative binomial (ZINB) models partially address excess zeros, their two-part formulation can still remain conservative around abrupt transitions.
To address these issues, we propose a model-agnostic learnable Tweedie head that can be attached as a plug-and-play output module to arbitrary ST-GNN backbones. 
Instead of likelihood-based Tweedie training, which typically requires surrogate objectives, our approach optimizes the closed-form Tweedie unit deviance and predicts the mean for point forecasting while learning a node-level variance power to capture heterogeneous variability across port areas.
Experiments on a maritime traffic graph constructed from real-world AIS data in the Port of Los Angeles and Long Beach show that the proposed head consistently improves RMSE across multiple ST-GNN backbones, especially on non-zero events, leading to more reliable forecasts for practical maritime traffic control.
\end{abstract}

\section{INTRODUCTION}

The rapid paradigm shift towards smart ports and intelligent transportation systems has intensified the demand for accurate vessel traffic flow prediction, which is essential for optimizing port logistics and improving navigational safety \cite{priya2024smart,wang2025ais,kim2019spatiotemporal}. However, forecasting maritime traffic remains highly challenging due to complex spatial dependencies and dynamic temporal fluctuations, 
which are widely recognized challenges in various spatio-temporal forecasting problems
\cite{liang2022fine,koo2024deep,choy2016looking,chung2019crime}.

\begin{figure}[thpb]
    \centering
    \includegraphics[width=3.2in]{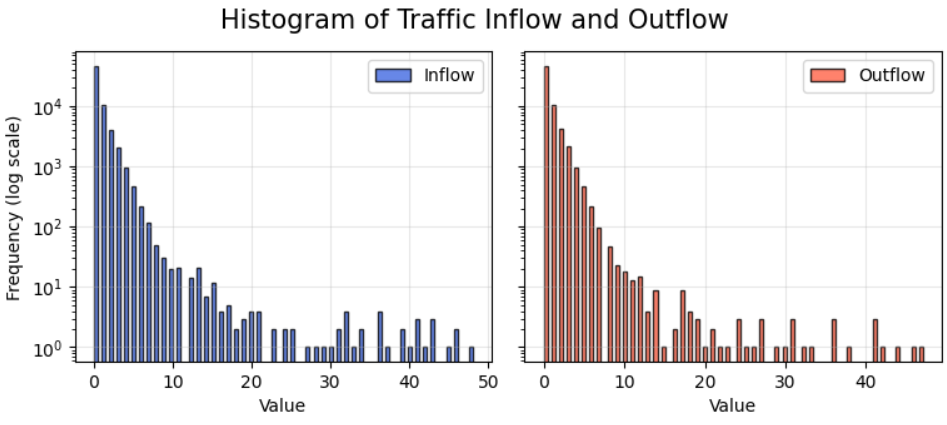}
    \caption{Empirical distributions of vessel traffic inflow and outflow derived from the real-world AIS dataset of the Los Angeles and Long Beach port region used in this study. Frequencies are plotted on a logarithmic scale. Both distributions exhibit severe zero inflation and long, heavy tails, reflecting sudden, bursty arrivals of vessels.}
    \label{fig:hist_log}
\end{figure}

While spatio-temporal graph neural networks (ST-GNNs) have achieved remarkable success in urban road networks \cite{yu2017spatio, li2017diffusion, wu2019graph}, robust forecasting in the maritime domain remains difficult. Marine areas are open spaces, leading to stochastic arrival patterns and highly volatile traffic volumes \cite{chen2025survey}. Unlike dense urban traffic, maritime movements exhibit inherent randomness and heterogeneity \cite{zhu2025multiport}, resulting in severe zero inflation and heavy-tailed sudden bursts, as illustrated in Fig. \ref{fig:hist_log}. In such scenarios, conventional ST-GNNs often experience significant performance degradation, tending to converge to near-zero predictions and failing to capture critical non-zero traffic spikes \cite{zhang2025uncertainty}.

To address data sparsity, recent studies have incorporated statistical models such as the Zero-Inflated Negative Binomial (ZINB) model \cite{zhu2025multiport, liang2024zinbgnn}. However, the traditional two-part mechanism of the ZINB model can suffer from poor predictive performance when its assumptions about zero generation do not align with the real-world data \cite{feng2021comparison}. Furthermore, the ZINB model struggles to adequately capture the long-tailed behavior associated with sudden traffic spikes \cite{jiang2023sttd}.

As an elegant alternative, the Tweedie distribution provides a unified framework for modeling zero outcomes and continuous positive values without requiring an explit zero-gating mechanism \cite{dunn2005series}. However, the application of Tweedie-based modeling in the maritime domain remains limited. Existing studies exploring Tweedie-based ST-GNNs in other domains \cite{jiang2023sttd, gao2024stzitdgnn} often rely on surrogate training objectives (e.g., lower-bound optimization) to bypass the intractable infinite series in the likelihood, which increases implementation complexity and may introduce approximation errors. 

To overcome these limitations, we propose a model-agnostic learnable Tweedie head. Instead of approximating the intractable infinite series required in likelihood-based methods, we optimize the Tweedie unit deviance, which provides a closed-form objective that ensures stable backpropagation without relying on lower-bound approximations. This formulation facilitates robust forecasting while allowing the model to dynamically estimate the node-level variance power and adaptively capture the spatial heterogeneity of individual port areas.

The main contributions of this paper are summarized as follows:

\begin{itemize}

\item 
We empirically show that conventional ST-GNNs suffer from significant performance degradation on highly sparse maritime traffic flow data 
and analyze the limitations of ZINB in capturing intermittent traffic spikes.

\item 
We propose a lightweight, learnable Tweedie head for vessel traffic flow prediction that can be attached to arbitrary ST-GNN backbones. It is optimized using the Tweedie unit deviance, providing a closed-form objective that avoids infinite-series likelihood approximations and enables stable training on zero-heavy, bursty traffic data.

\item 
We implement our module as a plug-and-play head across representative ST-GNN backbones. Extensive experiments on real-world AIS datasets from the Port of Los Angeles and Long Beach demonstrate that our framework significantly improves overall RMSE and performance on non-zero observations by accurately capturing sudden traffic spikes, providing a highly reliable signal for maritime traffic control.
\end{itemize}

\section{RELATED WORKS}

\subsection{Classical and Deep Learning Models for Vessel Traffic Flow Prediction}

Early vessel traffic forecasting relied on statistical time-series models such as ARMA and ARIMA \cite{wang2017arma, yoo2013arima}. While effective at capturing seasonal patterns, their linear structure oversimplifies the complex dynamics of vessel movements \cite{xu2022gru}. 
To address these nonlinear dynamics, deep learning models---including LSTMs, GRUs, and hybrid CNN-LSTM architectures---have been widely adopted \cite{xu2022gru, xie2018lstm, zhou2020lstmcnn}. LSTMs effectively capture complex temporal dependencies, while CNNs extract local spatial patterns from fixed grids \cite{kim2023contextual2,lee2023semi}. However, these methods encode space either implicitly or under Euclidean grid assumptions, which limits their ability to model the irregular, non-Euclidean topology of maritime networks \cite{zhu2025multiport,soh2018application,lee2013dependence}.

\subsection{Spatio-Temporal Graph Neural Networks}
Motivated by the success of ST-GNNs in urban road forecasting \cite{yu2017spatio,li2017diffusion,wu2019graph}, graph-based spatio-temporal models have been adopted in maritime settings to represent interactions among ports, waterways, or functional maritime regions \cite{chen2025survey}. Recent studies construct maritime graphs using data-driven connectivity---such as trajectory-derived feature points or historical port-to-port voyage frequencies---and apply ST-GNN models to learn the coupled spatio-temporal dependencies of vessel traffic volumes \cite{liang2022fine, mei2025mstgnn}.
Nevertheless, when the prediction target is zero-heavy and occasionally bursty, conventional regression-based training of ST-GNNs can lead to conservative near-zero forecasts and reduced sensitivity to sudden traffic spikes \cite{zhang2025uncertainty}.

\subsection{Zero-Inflated Models for Sparse Forecasting}

To handle sparsity in spatio-temporal forecasting, recent works integrate the ZINB distribution into spatio-temporal predictors. \cite{liang2024zinbgnn}  and \cite{zhuang2022stzinbgnn} introduced this approach for urban travel demand prediction, and \cite{zhu2025multiport} extended it to maritime origin-destination (O-D) forecasting. However, ZINB's two-part mechanism, which separately models structural zeros and count processes, can yield suboptimal results if the underlying zero-generation assumptions are violated \cite{feng2021comparison}. Additionally, these models often fail to capture the long-tail behavior of traffic spikes \cite{jiang2023sttd}.

\subsection{Tweedie Distribution as an Alternative}

As an alternative to two-part ZINB models, the Tweedie exponential dispersion family---specifically the compound Poisson-Gamma distribution---has been advocated for modeling zero-heavy, right-skewed data without explicit artificial zero-gates \cite{dunn2005series}. Building on these properties, recent studies have integrated the Tweedie distribution with ST-GNNs in specialized fields, such as ride-sharing O-D demand \cite{jiang2023sttd} and road-level traffic accident forecasting \cite{gao2024stzitdgnn}. However, because the exact Tweedie log-likelihood involves an intractable infinite series, these methods rely on surrogate training objectives. Specifically, they employ lower-bound approximations to bypass the summation formula during training. While effective for distribution calibration, these relaxations increase implementation complexity and computational overhead and may also introduce approximation errors.

In contrast, we propose a model-agnostic learnable Tweedie head for vessel traffic flow prediction, optimized via the Tweedie unit deviance. This closed-form objective avoids the intractable term in likelihood-based Tweedie training, enabling efficient and stable optimization. Moreover, our head learns the variance power $p$ for each node to capture spatial heterogeneity in maritime traffic and can be attached as a lightweight plug-and-play module to arbitrary ST-GNN backbones.

\section{METHODOLOGY}

\subsection{Problem Formulation}
We model the maritime traffic system as a directed graph $\mathcal{G}=(\mathcal{V},\mathcal{E},A)$, where $\mathcal{V}$ denotes a set of $N$ nodes (port and anchorage zones), $\mathcal{E}$ represents the set of edges
(navigational connections) and $A\in\mathbb{R}^{N\times N}$ is the weighted adjacency matrix. 
Given past observations $X\in\mathbb{R}^{T_{in}\times N\times C_{in}}$ (e.g., inflow/outflow), we forecast future traffic $Y\in\mathbb{R}^{T_{out}\times N\times C_{out}}$ using spatio-temporal representations produced by an ST-GNN backbone $f_{\text{ST-GNN}}(X,\mathcal{G})$.

\subsection{Baseline: ZINB-based Spatio-Temporal Forecasting}
As a sparsity-aware baseline for vessel traffic forecasting, we attach a ZINB head to the ST-GNN output. 
Given hidden features $H\in\mathbb{R}^{T_{out}\times N\times 3C_{out}}$, the head predicts $(\pi,\mu,\alpha)$ for each node and forecasting horizon, where $\pi$ is the zero-inflation probability, $\mu$ is the mean, and $\alpha$ is the dispersion parameter. 
ZINB models $Y$ using a two-part mixture and discounts the predictive mean as $\mathbb{E}[Y]=(1-\pi)\mu$.
The resulting predictive distribution is
\begin{equation}
P(Y=y)=\pi\mathbb{I}[y=0]+(1-\pi)\, P_{\text{NB}}(y;\mu,\alpha), \quad y\in\mathbb{N}_0,
\end{equation}
where $P_{\text{NB}}$ is the negative binomial probability mass function, and the model is trained by minimizing the negative log-likelihood.

\subsection{Proposed Method: Model-Agnostic Learnable Tweedie Head}

\begin{figure}[thpb]
    \centering
    \includegraphics[width=3.2in]{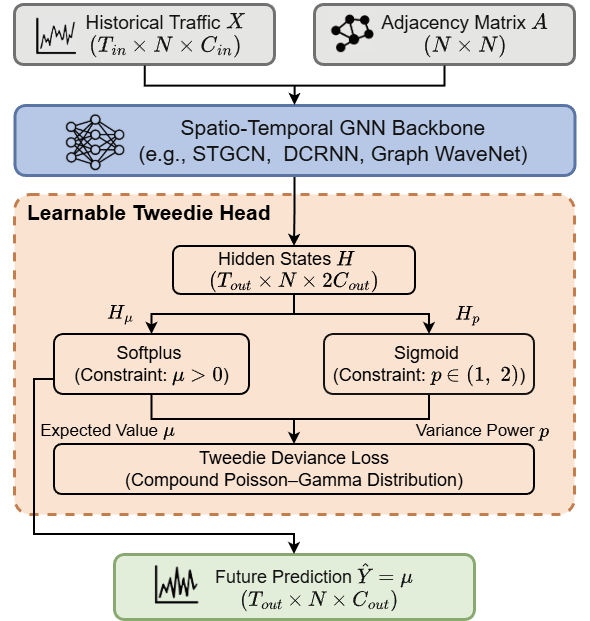}
    \caption{Overall architecture of the proposed ST-GNN forecasting framework with a learnable Tweedie head for vessel traffic flow prediction.}
    \label{fig:model}
\end{figure}

We propose a model-agnostic learnable Tweedie head that can be attached as a plug-and-play module to an arbitrary ST-GNN backbone (Fig.~\ref{fig:model}).

\subsubsection{Tweedie Modeling}
The Tweedie exponential dispersion family is characterized by the mean--variance relationship
$\mathrm{Var}(Y)=\phi\,\mu^{p}$, where $\mu>0$ is the mean, $\phi>0$ is the dispersion parameter, and $p$ is the variance power that controls the degree of overdispersion. 
When $1<p<2$, the Tweedie distribution corresponds to a compound Poisson--Gamma model, which supports an exact point mass at zero together with a right-skewed positive component, without requiring an explicit zero-gate mechanism as in ZINB.

\subsubsection{Head Parameterization}
Given backbone features $H=f_{\text{ST-GNN}}(X,\mathcal{G})\in\mathbb{R}^{T_{out}\times N\times 2C_{out}}$, we predict two node-wise parameters by splitting $H$ into $H_\mu$ and $H_p$:
\begin{equation}
\mu=\mathrm{Softplus}(H_\mu)+\epsilon,\qquad
p=1.01+0.98\,\sigma(H_p),
\end{equation}
where $\epsilon>0$ is a small constant. The node-level parameter $p$ enables the head to adaptively capture spatial heterogeneity across port areas, ranging from near-Poisson ($p \approx 1.1$) vessel traffic to overdispersed , Gamma-like ($p \approx 1.9$) traffic.

\subsubsection{Unit Deviance as a Training Objective}

For $1<p<2$, the Tweedie density can be written in an exponential dispersion form:
\begin{equation}
\log f_{\mathrm{TD}}(y\mid \mu,\phi,p)
=
\frac{1}{\phi}\left(
\frac{y\mu^{1-p}}{1-p}
-\frac{\mu^{2-p}}{2-p}
\right)
+\log a(y,\phi,p),
\label{eq:tweedie_ll}
\end{equation}
where $a(y,\phi,p)$ is a normalizing term. Likelihood-based training minimizes the negative log-likelihood
$\mathcal{L}_{\mathrm{NLL}}=-\sum_i \log f_{\mathrm{TD}}(y_i\mid \mu_i,\phi,p_i)$,
which requires evaluating $\log a(y,\phi,p)$. For $1<p<2$, this term is intractable and is typically expressed via an infinite-series summation. Accordingly, prior Tweedie ST-GNNs resort to surrogate objectives (e.g., lower-bound relaxations) to bypass the summation.

Instead, we optimize the Tweedie unit deviance (assuming $\phi=1$), which yields a closed-form loss and stable backpropagation.
For each observation $y\ge 0$, we use
\begin{equation}
\label{eq:tweedie_deviance}
d(y,\mu,p)=2\left(
\frac{y^{2-p}}{(1-p)(2-p)}
-\frac{y\,\mu^{1-p}}{1-p}
+\frac{\mu^{2-p}}{2-p}
\right),
\end{equation}
with the convention that the first term is $0$ when $y=0$. The batch loss is $\mathcal{L}_{\text{Tw}}=\frac{1}{|\mathcal{D}|}\sum_{i\in\mathcal{D}} d(y_i,\mu_i,p_i)$.

This objective is non-negative and equals zero if and only if $y = \mu$, providing a principled discrepancy for mean prediction while enforcing the Tweedie mean--variance relationship.

\subsubsection{Inference}
We use the Tweedie mean $\hat{y}=\mathbb{E}[Y]=\mu$ as the point forecast.

\section{EXPERIMENTS}

\subsection{Dataset and Graph Construction}
We constructed a high-resolution maritime traffic dataset from AIS records in the Los Angeles and Long Beach (LA/LB) region, spanning September 1 to November 30, 2024. 
To prevent look-ahead bias, the dataset was split chronologically into training, validation, and testing sets in a ratio of 70:10:20, corresponding to 1,528, 218, and 438 hours, respectively. All graph construction steps were performed using the training set only.

\subsubsection{Data-Driven Node Extraction} 
The AIS trajectories were spatially constrained to the LA/LB regions (Latitude: $33.50^\circ$ to $33.80^\circ$, Longitude: $-118.38^\circ$ to $-118.00^\circ$). To identify the port and anchorage zones, near-stationary records (Speed Over Ground (SOG) $\le 1.0$ knot) were filtered and clustered using the DBSCAN algorithm. Considering distinct operational characteristics, terminals and anchorages were separated based on a breakwater boundary, with domain-specific hyperparameters applied to each: $\epsilon=250$\,m for terminals and $\epsilon=1000$\,m for anchorages, while $\text{Minimum Points}=400$ was maintained for both. Each cluster was subsequently converted to a polygon boundary using an alpha-shape procedure, with radius thresholds of  400\,m for terminals and 2{,}000\,m for anchorages. This process resulted in the extraction of $|\mathcal{V}| = 30$ polygonal nodes, capturing the irregular shapes of port functional areas (shown as red and blue polygons in Fig. \ref{fig:heatmap}).

\subsubsection{Edge Construction and Weighted Adjacency Matrix}
\begin{figure}[thpb]
    \centering
    \includegraphics[width=3.2in]{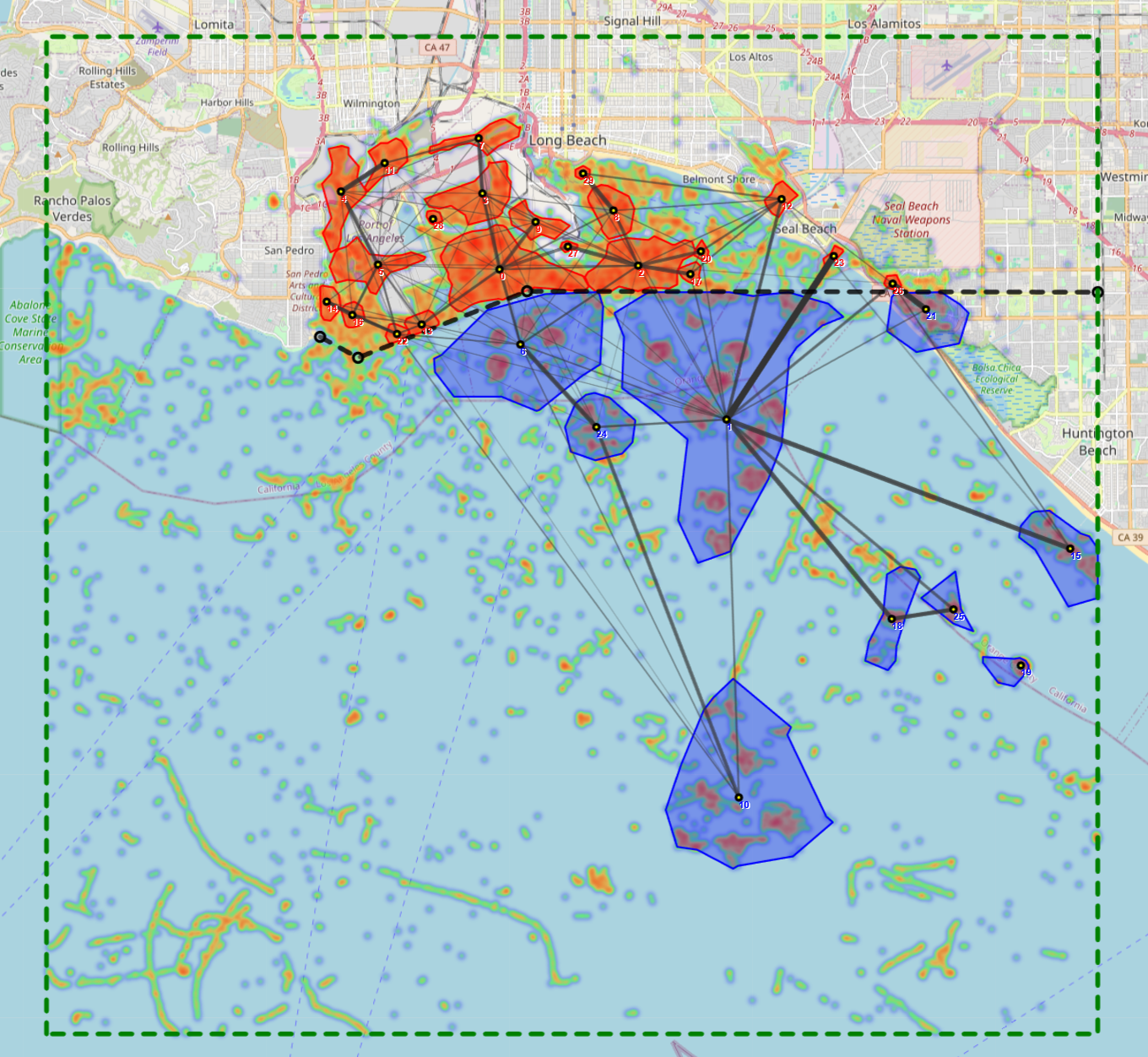}
    \caption{Visualization of the constructed graph for the LA/LB port. The background heatmap illustrates the spatial density of stationary AIS signals. The green and black dashed lines represent the study area and the breakwater boundary, respectively. Red polygons denote terminal nodes inside the breakwater, while blue polygons represent anchorage nodes outside. Edge thickness is proportional to the transition probability (weight).}
    \label{fig:heatmap}
\end{figure}

We constructed directed edges $\mathcal{E}$ by tracking the sequential transitions of individual vessels between nodes. To filter out noise, we only retained edges where the transition count reached a threshold of at least 10 vessels.
The weighted adjacency matrix $A$ was constructed via row-wise normalization, where each element $A_{i,j}$ denotes the ratio of vessel transition counts from node $i$ to node $j$ relative to the total departures from node $i$.
This resulted in 198 directed edges with an average transition weight of $0.1414$, as visualized in Fig. \ref{fig:heatmap}.

\subsubsection{Traffic Flow Aggregation}
We tracked vessel trajectories to count the hourly number of vessels entering (inflow) and exiting (outflow) each polygonal node. This generated the spatiotemporal dataset $\mathcal{X} \in \mathbb{R}^{T \times 30 \times 2}$, where $T$ is 1,528, 218, and 438 for the training, validation, and testing sets, respectively.

\begin{table*}[t]
\centering
\caption{Vessel traffic flow prediction performance of baseline ST-GNNs and their sparsity-aware variants. The best results for each backbone are highlighted in bold.}
\label{tab:performance_benchmark}
\resizebox{\textwidth}{!}{
\begin{tabular}{l ccc ccc ccc}
\toprule
\multirow{2}{*}{\textbf{Metric}} & \multicolumn{3}{c}{\textbf{GWNet}} & \multicolumn{3}{c}{\textbf{DCRNN}} & \multicolumn{3}{c}{\textbf{STGCN}} \\
\cmidrule(lr){2-4} \cmidrule(lr){5-7} \cmidrule(lr){8-10}
& Base & +ZINB & +Tweedie (Ours) & Base & +ZINB & +Tweedie (Ours) & Base & +ZINB & +Tweedie (Ours) \\
\midrule
MAE (All) & \textbf{0.3970} & 0.4773 & 0.4873 & \textbf{0.3986} & 0.4823 & 0.4735 & \textbf{0.3905} & 0.4907 & 0.4827 \\
MAE (Non-zero) & 1.1969 & 0.9940 & \textbf{0.9709} & 1.2530 & \textbf{0.9681} & 0.9794 & 1.1902 & \textbf{0.9623} & 0.9723 \\
RMSE (All) & 0.9172 & 0.9112 & \textbf{0.8951} & 0.9255 & 0.8971 & \textbf{0.8802} & 0.9024 & 0.9164 & \textbf{0.8865} \\
RMSE (Non-zero) & 1.6720 & 1.5241 & \textbf{1.4570} & 1.7432 & 1.5089 & \textbf{1.4831} & 1.6682 & 1.4877 & \textbf{1.4634} \\
\bottomrule
\end{tabular}
}
\end{table*}

\subsection{Baseline Models}
To validate the performance and the model-agnostic nature of the proposed Tweedie head, we evaluate it on three widely used ST-GNN backbones that represent distinct design choices:
\begin{itemize}
    \item STGCN: Purely spatio-temporal graph convolutions for efficient feature extraction  \cite{yu2017spatio}.
    \item DCRNN: Diffusion graph convolution with seq2seq GRU to model directional dynamics \cite{li2017diffusion}.
    \item Graph WaveNet (GWNet): Dilated causal temporal convolutions with adaptive graph learning for long-range dependencies  \cite{wu2019graph}.
\end{itemize}

For each backbone, we compare three output-head variants: 
(i) Base, trained with mean absolute error (MAE); 
(ii) +ZINB, a ZINB head as a representative baseline for zero-heavy targets; and (iii) +Tweedie (Ours), Tweedie head optimized via unit deviance for spike-sensitive point forecasting.

We use the same forecasting setup for all variants: an input window of 24 hours and a prediction horizon of 6 hours. 
All models are trained for 100 epochs with Adam (learning rate 0.001), and the backbone hidden dimension is fixed to 32 across architectures.

\subsection{Evaluation Metrics}
We report mean absolute error (MAE) and root mean square error (RMSE) as the standard evaluation metrics. However, due to the sparsity of maritime traffic data (over 70\% zero values), overall scores alone can be misleading. A trivial near-zero predictor may achieve low MAE while failing to capture actual arrivals.

To better assess the ability to forecast sudden traffic spikes, we additionally report Non-zero MAE and Non-zero RMSE, computed only on timesteps with positive ground-truth values. These metrics better reflect the model's ability to capture real traffic activity by mitigating the bias introduced by frequent zeros.

\begin{figure*}[t]  
    \centering
    \includegraphics[width=\textwidth]{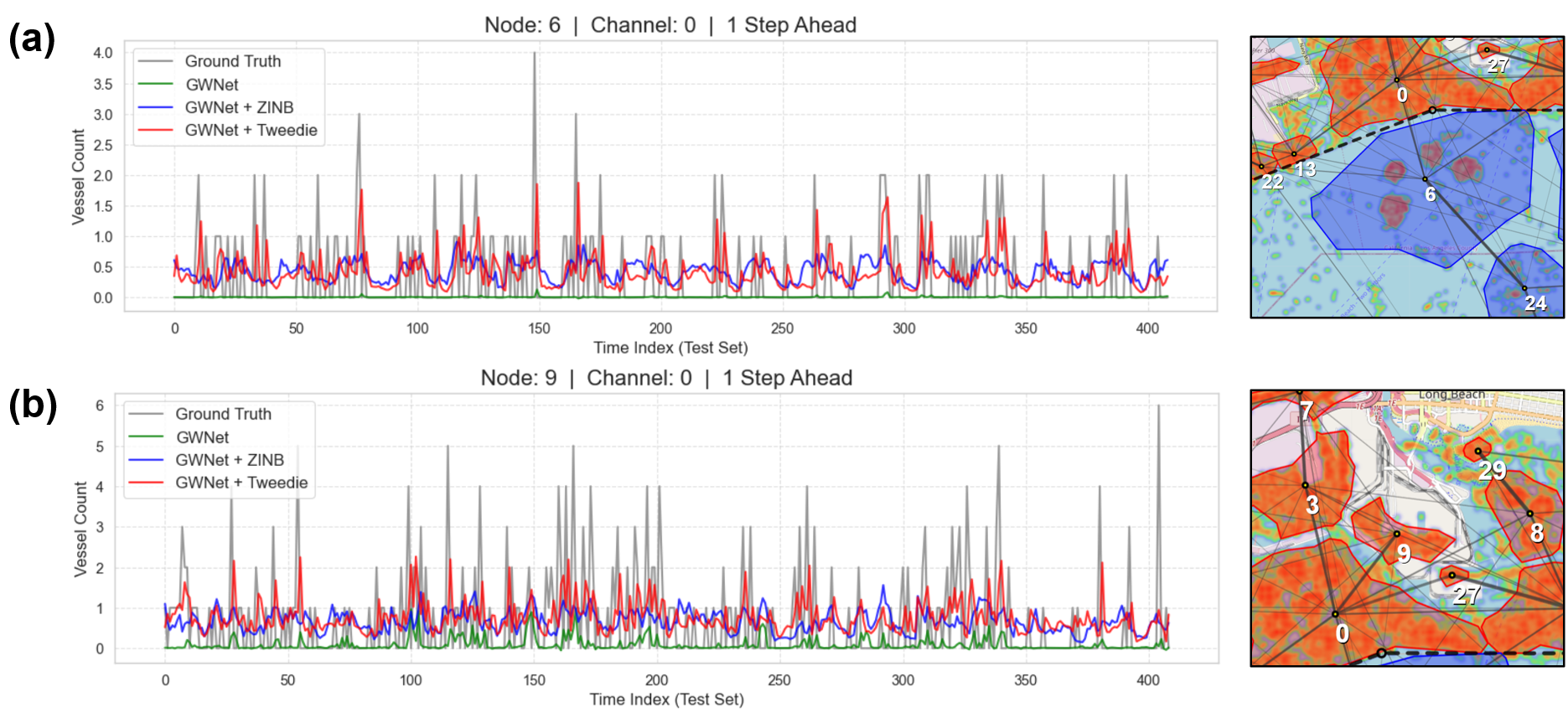} 
    \caption{Visualization of 1-step-ahead forecasting results on the test set for two representative nodes. 
For each subfigure, the left panel shows the vessel-count time series (ground truth in gray) and the corresponding predictions from GWNet (green), GWNet+ZINB (blue), and GWNet+Tweedie (red), while the right panel visualizes the geographic location of the polygonal node in the LA/LB graph. 
(a) Node 6 (anchorage zone outside the breakwater). (b) Node 9 (inner-port terminal zone).}
    \label{fig:case_st}
\end{figure*}

\subsection{Quantitative Results}
The comprehensive prediction performances of the 9 model combinations are summarized in Table \ref{tab:performance_benchmark}. We highlight three observations:

\subsubsection{Misleading Overall Metrics on Zero-Heavy Targets}
The Base models achieve the best overall MAE (e.g., 0.3970 for GWNet), yet their errors on non-zero events are substantially larger. This indicates that conventional ST-GNNs can bias predictions toward near-zero values and miss important vessel traffic flow.

\subsubsection{Improved Non-Zero Accuracy with Sparsity-Aware Heads}
Adding sparsity-aware output heads (+ZINB and +Tweedie) consistently reduces Non-zero MAE across backbones, reflecting improved sensitivity to non-zero traffic events. 
ZINB and Tweedie achieve comparable Non-zero MAE, with small backbone-dependent differences.

\subsubsection{RMSE Improvements with Tweedie}
The most consistent gains of our method appear in RMSE, particularly in Non-zero RMSE. 
Across all three backbones (STGCN, DCRNN, and GWNet), the +Tweedie variants achieve the lowest overall RMSE and Non-zero RMSE. 
Since RMSE penalizes large errors more heavily than MAE, these results indicate that the Tweedie head better captures the magnitude of sudden traffic surges than the baselines. 
These improvements are consistent across STGCN, DCRNN, and GWNet, supporting the model-agnostic nature of the proposed head.

\subsection{Qualitative Analysis and Case Study}

We further examine 1-step-ahead forecasts to visualize model behavior under severe zero-inflation and intermittent surges.
We select two representative nodes with distinct operational contexts: Node 6 (anchorage zone outside the breakwater) and Node 9 (highly active inner-port terminal zone).

\subsubsection{Near-Zero Predictions of Baseline}
As observed in Fig. \ref{fig:case_st}, the GWNet baseline largely fails to capture the dynamic fluctuations of the actual vessel traffic flow. It tends to produce flatlined or conservative near-zero forecasts, which may reduce  overall error but fail to track non-zero surges in the ground truth. 
This behavior is consistent with the performance degradation of  conventional ST-GNNs  under highly zero-inflated targets \cite{ zhang2025uncertainty, rovzanec2025dealing}.

\subsubsection{ZINB and Gate-Discounted Mean}
GWNet+ZINB improves sensitivity to non-zero events compared to the Base GWNet, but it can still underestimate peak magnitudes. For instance, when the ground-truth traffic experiences a spike of 4 to 5 vessels in Fig. \ref{fig:case_st} (b), the ZINB model's prediction barely reaches 1 vessel.
A plausible explanation is that the predictive mean is discounted as $\hat{y}=(1-\pi)\mu$; if the zero-inflation gate $\pi$ does not decrease quickly during a transition, the expected value remains conservative.

\subsubsection{Enhanced Spike Predictions of Tweedie Head}
In contrast, our proposed GWNet+Tweedie model (red line in Fig. \ref{fig:case_st}) exhibits significantly sharper and more proportional predictions for both nodes. While perfectly matching highly stochastic bursts remains challenging, the Tweedie head more closely tracks the rise and fall patterns in both the terminal and anchorage nodes.

\section{Conclusion}

This paper addressed node-level vessel traffic flow prediction in the LA/LB port region, where the target data are highly zero-inflated with intermittent bursts. 
We showed that conventional ST-GNN backbones tend to converge to conservative near-zero forecasts, yielding deceptively low overall errors while missing non-zero traffic activity. 
To mitigate this problem, we introduced a model-agnostic learnable Tweedie head that can be attached as a plug-and-play output module to arbitrary ST-GNN backbones. 
Across three representative architectures (STGCN, DCRNN, and Graph WaveNet), the proposed head consistently achieved the best RMSE and Non-zero RMSE, and improved accuracy in predicting traffic spikes compared to base ST-GNN models and ZINB variants. 
These results indicate that augmenting existing ST-GNNs with the proposed Tweedie head can substantially improve practical forecasting performance in zero-heavy maritime traffic settings.

Despite these consistent gains, the model still tends to remain conservative during the most extreme events; accurately matching the absolute maxima of traffic surges remains challenging in our setting.
Our evaluation is also limited to a single region (LA/LB) and a three-month period with hourly aggregation, and generalization across ports, seasons, and temporal resolutions requires further validation.
Finally, the current framework focuses on point forecasting and uses a transition-based static graph; external drivers (e.g., weather, schedules) and time-varying connectivity are not explicitly modeled \cite{kim2023contextual}.

Future work includes cross-region evaluation and transfer to other ports and waterways, incorporating exogenous information and dynamic graph updates, and exploring tail-aware objectives or hybrid approaches to better handle extreme traffic spikes. 
Extending the head toward calibrated uncertainty estimation is another promising direction for operational decision support \cite{kim2021locally,yoon2024uncertainty, yoon2026uncertainty}.

\section*{Acknowledgment}
This work was supported by the National Research Foundation of Korea (NRF) grant funded by the Korea government (MSIT) (2023R1A2C2005453, RS-2023-00218913).










\bibliographystyle{IEEEtran}  
\bibliography{my_refs}

\end{document}